\documentclass{ecai} 
\usepackage{latexsym}
\usepackage{amssymb}
\usepackage{amsmath}
\usepackage{amsthm}
\usepackage{booktabs}
\usepackage{enumitem}
\usepackage{graphicx}
\usepackage{color}

\newcommand{\BibTeX}{B\kern-.05em{\sc i\kern-.025em b}\kern-.08em\TeX}

\begin{document}

%%%%%%%%%%%%%%%%%%%%%%%%%%%%%%%%%%%%%%%%%%%%%%%%%%%%%%%%%%%%%%%%%%%%%%%%

\begin{frontmatter}

%%% Use this command to specify your submission number.
%%% In doubleblind mode, it will be printed on the first page.

\paperid{561} 

%%% Use this command to specify the title of your paper.

\title{FsPONER: Few-shot Prompt Optimization for Named Entity Recognition in Domain-specific Scenarios}
%\title{Named Entity Recognition Using Instruction-following Large Language Models for Industrial Domains}

%%% Use this combinations of commands to specify all authors of your 
%%% paper. Use \fnms{} and \snm{} to indicate everyone's first names 
%%% and surname. This will help the publisher with indexing the 
%%% proceedings. Please use a reasonable approximation in case your 
%%% name does not neatly split into "first names" and "surname".
%%% Specifying your ORCID digital identifier is optional. 
%%% Use the \thanks{} command to indicate one or more corresponding 
%%% authors and their email address(es). If so desired, you can specify
%%% author contributions using the \footnote{} command.

% \author[A]{\fnms{First}~\snm{Author}\orcid{....-....-....-....}\thanks{Corresponding Author. Email: somename@university.edu.}\footnote{Equal contribution.}}
% \author[B]{\fnms{Second}~\snm{Author}\orcid{....-....-....-....}\footnotemark}
% \author[B,C]{\fnms{Third}~\snm{Author}\orcid{....-....-....-....}} 

% \address[A]{Short Affiliation of First Author}
% \address[B]{Short Affiliation of Second Author and Third Author}
% \address[C]{Short Alternate Affiliation of Third Author}
\author[A,B]{\fnms{Yongjian}~\snm{Tang}\thanks{Yongjian Tang. Email: yongjian.tang@tum.de.}%\footnote{Equal contribution.}
}
\author[B]{\fnms{Rakebul}~\snm{Hasan}}
\author[A,B]{\fnms{Thomas}~\snm{Runkler}
} 

\address[A]{Technical University of Munich, Germany}
\address[B]{Siemens AG, Munich, Germany}

%%% Use this environment to include an abstract of your paper.
\begin{abstract}
%Large Language Models (LLMs) have provided a new pathway for Named Entity Recognition (NER) tasks. Compared with fine-tuning, LLM-based prompting methods avoid the need for training, conserve substantial computational resources, and rely on minimal annotated data. Previous studies have achieved comparable performance to fully-supervised BERT-based fine-tuning approaches on general NER benchmarks. However, none of the previous approaches has investigated the efficiency of prompting in industrial domains. To address this gap, we introduce FsPONER, a novel few-shot prompt optimization approach for NER, and evaluate its performance on three industrial NER datasets while using multiple LLMs, including GPT-4-32K, GPT-3.5-Turbo, LLaMA 2-chat, and Vicuna. FsPONER consists of three few-shot selection methods based on random sampling, sentence embedding, and TF-IDF vectors. We investigate these different methods as the number of few-shot examples increases and compare their optimal performance with fine-tuned BERT and LLaMA 2-chat 7B. In the considered real-world industrial scenarios with data scarcity, FsPONER surpasses fine-tuned models by approximately 10\% in F1 score.
Large Language Models (LLMs) have provided a new pathway for Named Entity Recognition (NER) tasks. Compared with fine-tuning, LLM-powered prompting methods avoid the need for training, conserve substantial computational resources, and rely on minimal annotated data. Previous studies have achieved comparable performance to fully supervised BERT-based fine-tuning approaches on general NER benchmarks. However, none of the previous approaches has investigated the efficiency of LLM-based few-shot learning in domain-specific scenarios. To address this gap, we introduce FsPONER, %a novel few-shot prompt optimization approach for NER, and evaluate its performance on domain-specific datasets and take industrial manufacturing and maintenance as examples
a novel approach for optimizing few-shot prompts, and evaluate its performance on domain-specific NER datasets, with a focus on industrial manufacturing and maintenance, while using multiple LLMs -- GPT-4-32K, GPT-3.5-Turbo, LLaMA 2-chat, and Vicuna. %FsPONER consists of three few-shot selection methods based on random sampling, TF-IDF vectors, and a combination of both. We investigate these different methods as the number of few-shot examples increases and compare their optimal NER performance with fine-tuned BERT and LLaMA 2-chat 7B. In the considered real-world scenarios with data scarcity, FsPONER surpasses fine-tuned models by approximately 10\% in F1 score.
FsPONER consists of three few-shot selection methods based on random sampling, TF-IDF vectors, and a combination of both. We compare these methods with a general-purpose GPT-NER method as the number of few-shot examples increases and evaluate their optimal NER performance against fine-tuned BERT and LLaMA 2-chat. In the considered real-world scenarios with data scarcity, FsPONER with TF-IDF surpasses fine-tuned models by approximately 10\% in F1 score.

\end{abstract}

\end{frontmatter}

%%%%%%%%%%%%%%%%%%%%%%%%%%%%%%%%%%%%%%%%%%%%%%%%%%%%%%%%%%%%%%%%%%%%%%%%
\section{Introduction}
%\subsection{Named Entity Recognition (NER)}
Named Entity Recognition (NER) is a common information extraction task, in which we identify and categorize the key information in the text. The strategies to solve such sequence labeling tasks have been evolving over time. NER systems were initially crafted with semantic and syntactic rules to recognize entities~\cite{etzioni2005unsupervised, sekine2004definition}. Due to the domain-specific rules and incomplete dictionaries, such rule-based NER systems cannot be transferred to other domains. Another strategy comprises unsupervised NER systems~\cite{collins1999unsupervised, nadeau2006unsupervised}, which detect entities from clustered groups exhibiting similar contextual information and deduce the correct entities based on the statistical lexical patterns computed on a large corpus. Furthermore, with the advent of deep neural networks, NER systems were empowered to learn distributed word representations and complex features from raw data automatically~\cite{lample2016neural}. Facilitated by the transformer architecture~\cite{vaswani2017attention}, %a considerable number of language models, e.g.~BERT, were trained to learn context-aware representations. With an encoder structure, BERT stands out as a highly adopted model for NER tasks. 
such pre-trained language models have advanced the state-of-the-art NER performance progressively over the past decade~\cite{devlin2019bert, liu2019roberta}. Nevertheless, either training a model from scratch or fine-tuning an existing model requires extensive data tailored to a specific domain and this can be highly cost-intensive in real-world applications. 

\begin{figure}[h!]
\centering
\includegraphics[width=5.5cm]{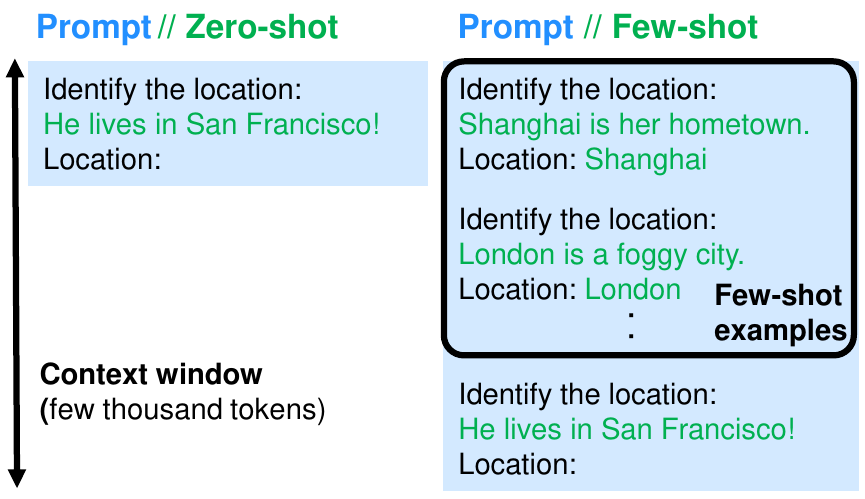}
\caption{A NER example using zero-shot and few-shot prompting.}
\vspace{0.5cm}
\label{fig:fewshot_intro}
\end{figure}

In recent years, Foundation LLMs, such as GPT models~\cite{radford2019language, openai2023gpt4, brown2020language}, have demonstrated exceptional competence in knowledge inference and information extraction, offering us an alternative solution to NER. Their instruction-following abilities~\cite{zhang2023instruction, ouyang2022training} facilitate a more streamlined and accessible problem-solving procedure, while simultaneously reducing the need for massive annotated data. Fig.~\ref{fig:fewshot_intro}~illustrates an example of leveraging LLMs to solve a NER task. In a zero-shot setting, the LLM identifies the named entities relying on the prior knowledge obtained in pre-training. The performance can be enhanced by incorporating a sequence of few-shot examples into the prompt. These examples serve as demonstrations, allowing the LLM to glean information of the presented domain and refine its understanding, leading to a more precise extraction of entity types. 

In previous studies, LLM-based prompting has achieved comparable performance to fully supervised baselines on standard NER benchmarks~\cite{wang2023gpt, xie2023empirical, xu2023large}, but none of these studies evaluate the performance of LLMs on domain-specific use cases. Moreover, these studies exclusively compare the LLMs using prompting techniques with middle-sized pre-trained language models, such as fine-tuned BERT variants~\cite{devlin2019bert, liu2019roberta, sanh2019distilbert}, and a comparison to fine-tuned LLMs of the same size is still missing. To the best of our knowledge, we are the first to include both genuine LLMs using few-shot prompting and instruction fine-tuned LLMs in experiments. The contribution of this work can be summarized as follows:
\vspace{-0.1cm}
\begin{itemize}
    \item [(i)] We propose FsPONER with three variants with different few-shot selection methods to filter semantically close examples and optimize the prompt.
    \item [(ii)] We investigate how the number of few-shot examples and the scale of few-shot datasets influence the NER performance.  
    %\item [(iii)] We evaluate FsPONER on domain-specific NER scenarios, exemplified by the evaluation results in industrial manufacturing and maintenance, while leveraging four LLMs -- GPT-4-32K, GPT-3.5-Turbo, LLaMA 2-chat, and Vicuna.
    \item [(iii)] We exemplify the evaluation results of FsPONER on industrial manufacturing and maintenance use cases to assess its effectiveness in domain-specific NER scenarios, using four LLMs – GPT-4-32K, GPT-3.5-Turbo, LLaMA 2-chat, and Vicuna.
    %\item [(iv)] We compare the LLMs using FsPONER with fine-tuned LLaMA 2-chat 7B and BERT, observing that FsPONER attains a 10\% higher F1 score in the considered scenarios with data scarcity. %analyzing the advantages of both fine-tuning and prompting in the considered industrial scenarios.
    \item [(iv)] We compare FsPONER with a non-domain-specific GPT-NER method and evaluate their optimal NER performance against fine-tuned LLaMA 2-chat 7B and BERT, observing that FsPONER attains a 10\% higher F1 score in the considered scenarios with data scarcity. 
\end{itemize} 

% alternative to semantic few-shot examples
% include GPT-4, and more open-source models
% evaluate on domain-specific use cases
%\textbf{Plateaued performance of LLMs}
%\textbf{No domain-specific evaluation}

\section{Related Work}
\label{related_work}
\subsection{Development of Language Models}
Language modeling~\cite{zhao2023survey} stands out as a core technique to advance language intelligence of machines and has received extensive attention over the past decade. The story begins with statistical language models~\cite{jelinek1998statistical}, which follow the Markov assumption and predict the next word based on the most recent context. %Due to the foundation in statistical learning methods, SLMs find wide applications in information retrieval tasks~\cite{zhai2008statistical}. 
The rise of deep neural architectures opens the stage for neural language models. As a milestone, Bengio et al.~\cite{bengio2000neural} introduced the concept of distributed word representations and crafted word prediction functions on aggregated context features. Furthermore, Collobert et al.~proposed a unified multi-layer neural architecture by discovering the internal representations of unlabeled datasets~\cite{collobert2011natural}, while Mikolov et al.~\cite{mikolov2013distributed, mikolov2013efficient} proposed word2vec, a simplified shallow neural network designed for learning word representations effectively. These studies have initiated the utilization of language models for representation learning, elevating word sequence modeling to a more advanced level.

Vaswani et al.~\cite{vaswani2017attention} proposed the transformer architecture and attention mechanism, which started the generation of pre-trained language models.
In alignment with this highly parallelizable architecture, language models were trained to learn context-aware word representations. BERT, a model proposed by Devlin et al.~\cite{devlin2019bert} and pre-trained on large-scale unlabeled corpora bidirectionally, stands out. %as a highly adopted model. 
The semantic representations obtained in pre-training make BERT approachable for a broad spectrum of downstream tasks in specific domains. %~\cite{shrivastava2022named, beltagy2019scibert}. 
Inspired by such ”pre-training” and ”fine-tuning” modes, a substantial quantity of follow-up works have been developed over time, e.g.~RoBERTa~\cite{liu2019roberta} and DistilBERT\cite{sanh2019distilbert}. 

The research community continues to enhance the performance of language models by scaling up their sizes. Compared with their smaller counterparts, large-scale models demonstrate unseen emergent abilities~\cite{wei2022emergent, lu2023emergent} in solving complex tasks, which have provided us with more information-based problem-solving possibilities, such as in-context learning%~\cite{lu2023emergent, dong2022survey, wei2023larger, liu2023lost, rubin2021learning}
, prompt engineering, step-by-step reasoning 
%~\cite{wei2022chain, kojima2022large, besta2024graph}
and retrieval augmented generation. %~\cite{lewis2020retrieval, cai2022recent, jiang2023active}
%and prompt engineering%~\cite{lu2021fantastically, lou2023prompt, rubin2021learning}
These techniques have enlightened this work, leveraging few-shot prompting to solve domain-specific NER tasks.

\subsection{In-context Learning}
In-context learning~\cite{dong2022survey} refers to the process of learning from analogy, where a query and a piece of demonstration context are merged into a prompt and then fed into LLMs to obtain the required outcomes. In contrast to supervised learning, in which model parameters are updated at training or fine-tuning stages, ICL does not perform any parameter updates, but generates the answer directly based on the provided information. The prompt serves as an activator, enabling LLMs to understand the critical information within the demonstration context comprehensively.

Wei et al.~\cite{wei2023larger} illustrate that LLMs can override semantic pre-trained knowledge when presented with conflicting in-context examples. The ability to surpass semantic priors enhances as the model size increases. Smaller models cannot flip predictions and follow contradictory labels, while larger models can perform this effectively. 
Liu et al.~\cite{liu2023lost} utilize LLMs to retrieve relevant information within a long context. They find that the performance is optimal when relevant information occurs at the beginning or end of the input context, and it degrades significantly when models must access relevant information in the middle. Moreover, as the input context grows longer, even explicitly designed long-context models fail to identify the relevant information effectively.

Many creative ideas of ICL have been proposed. 
Li et al.~\cite{li2023emotionprompt} introduce EmotionPrompt, which incorporates emotional stimulus into prompts and emphasizes the task significance to improve the performance of LLMs. Ye et al.~\cite{ye2023investigating} add an instruction set to the prompt and enhance the zero-shot generalization abilities of LLMs. Wei et al.~\cite{wei2023symbol} present symbol tuning. They replace in-context input–label pairs with arbitrary symbols and find that symbol-tuned LLMs are better at ICL than original models, especially in settings where relevant labels are not available.

In terms of reasoning, Kojima et al.~\cite{kojima2022large} show that LLMs are decent zero-shot reasoners by adding an instructive command at the end of the prompt template. Wei et al.~\cite{wei2022chain} propose chain-of-thought prompting and improve the performance of LLMs on a range of arithmetic, commonsense, and symbolic reasoning tasks. Built on this, Yao et al.~\cite{yao2023tree} introduce tree-of-thought, which allows language models to perform deliberate decision-making by considering multiple different reasoning paths and self-evaluating choices to decide the next course of action. Moreover, Besta et al.~\cite{besta2024graph} propose graph of thoughts, in which the information generated by LLMs is modeled as an arbitrary graph,  where units of information %(“LLMs' thoughts”) 
are vertices, and edges correspond to dependencies between these vertices.

\subsection{In-context Learning with Few-shot Examples}
%Few-shot learning is a machine learning paradigm where models are enabled to accomplish a task with only a small number of training data. 
The in-context learning ability of LLMs has facilitated a new form of few-shot learning -- few-shot prompting, in which demonstration examples are integrated into the context window directly. %, as showed in Figure \ref{fig:fewshot_intro}.

Lu et al.~\cite{lu2021fantastically} find that the order in which few-shot examples are permutated in the prompt can make the difference between near state-of-the-art and random guess performance. %To explore the connection between sample order sensitivity and different model sizes, they choose a fixed random subset of four samples with a balanced label distribution %from the SST-2 dataset and consider all 24 possible sample order permutations. 
Larger models generally achieve better performance with low variance in the experiments, and adding more few-shot samples into prompts does yield a noticeable enhancement in performance, but it does not significantly reduce variance. To find the optimal sample permutations, they propose the idea of ordering few-shot examples based on entropy, which yields a 13$\%$ relative improvement for GPT family models across eleven different text classification tasks.

%proposed the idea of selecting few-shot examples in the embedding space of a sentence encoder. They leverage RoBERTa-large~\cite{liu2019roberta} to transform original sentences into embedding vectors. Based on the obtained embedding vectors, they apply the $k$NN algorithm to find the K nearest examples for the input sentence and use them as few-shot examples. 

Furthermore, Liu et al.~\cite{liu2021makes} propose %propose to retrieve examples that are semantically similar to a test sample to formulate its corresponding prompt. 
the idea of selecting few-shot examples in the embedding space of a sentence encoder. They leverage RoBERTa-large~\cite{liu2019roberta} to transform original sentences into embedding vectors. Based on these vectors, they apply the $k$NN algorithm to identify the \(k\) nearest examples for the input sentence and use them as few-shot examples. The selected in-context examples can provide more informative inputs to unleash GPT-3’s extensive knowledge. Such refined few-shot selection strategy has shown efficiency in advancing the performance of LLMs across a wide range of NLP tasks, including question answering, machine translation, and information extraction. 

%If no labeled data sample is given, the problem is simplified to the zero-shot setting, where pre-trained models predict the class labels of unknown data without any additional information. Kojima et al.~\cite{kojima2022large} show that LLMs are decent zero-shot reasoners by simply adding an instructive command at the end of the prompt template. 
%These methods bring LLM reasoning closer to human thinking processes and advance the research on ICL learning consistently. 

\subsection{Few-shot In-context Learning for NER}
Regarding the application of LLMs in NER tasks, Wang et al.~\cite{wang2023gpt} integrate semantically close examples into the prompt and analyze the impact of few-shot quantity and quality on NER performance. They enhance the embedding-based few-shot selection method~\cite{liu2021makes} by replacing RoBERTa-large~\cite{liu2019roberta} with SimCSE~\cite{gao2022simcse}, a contrastive learning framework developed for sentence embedding, and using it to encode the few-shot examples and input sentences. They name the approach GPT-NER, which exhibits efficiency in low-resource and few-shot setups, and has attained comparable performance to fully-supervised BERT-based approaches on two general NER benchmarks. 

Xie et al.~\cite{xie2023empirical} investigate the zero-shot performance of ChatGPT~\cite{radford2019language} in information extraction, in which they adapt reasoning methods and decompose the NER task into multiple simpler sub-problems. Their method enhances the zero-shot NER performance across seven benchmarks, including Chinese and English datasets as well as domain-specific and general-purpose scenarios. However, their study does not encompass few-shot learning.

%\subsubsection{Zero-shot learning}
%\subsection{Selection of Few-shot Examples}
%Existing works that try to select few-shot examples with close semantic embedding.
\section{FsPONER}
\begin{figure}[h!]
    \vspace{-0cm}
    \includegraphics[width=0.97\linewidth]{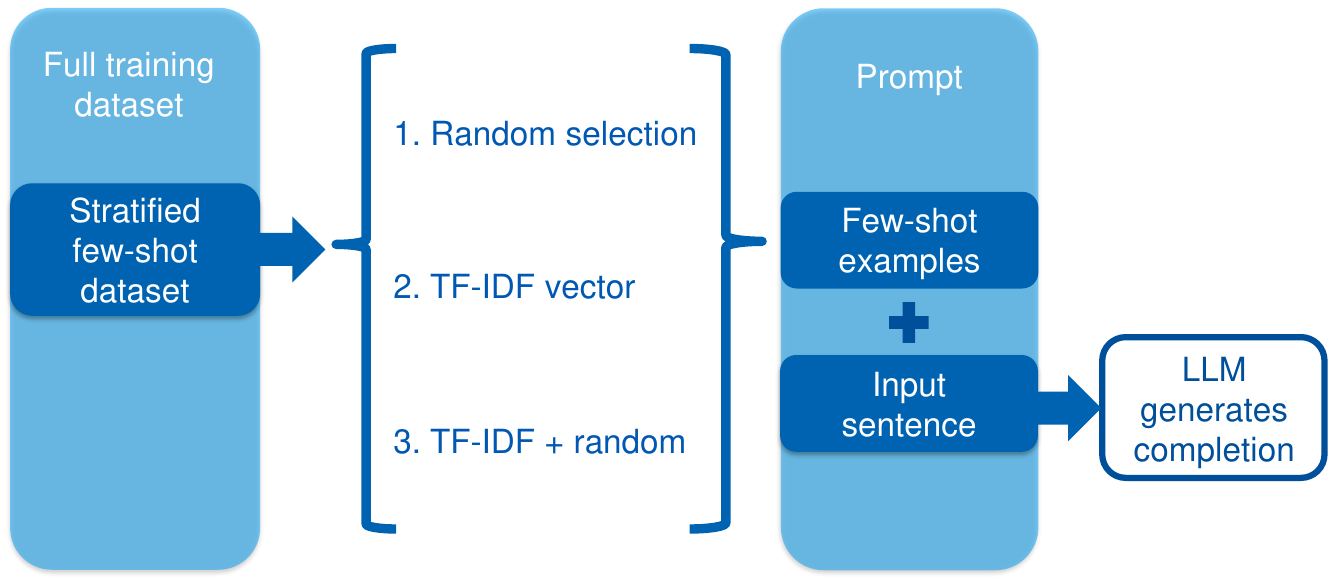}
    \centering
    \caption{Overview of FsPONER.}
    \vspace{0.2cm}
    \label{3_methods}
\end{figure}

Previous studies have focused on identifying semantically similar examples to the input in the embedding space, which overlooked the entity distribution within the dataset. When dealing with infrequent entity types in specific domains, LLMs may struggle to accurately extract entities using the universal prior knowledge acquired during pre-training. To address this, we introduce data stratification as a preliminary step to consider all entity types fairly and propose FsPONER, a LLM-based NER framework with three different few-shot selection methods: random sampling, TF-IDF vectors, and a combination of both. The goal is to incorporate term frequency alongside sentence embedding, enhancing domain-specific NER performance. 
We follow the outline in Fig.~\ref{3_methods} and demonstrate the principle of prompt optimization in FsPONER.

\subsection{Stratified Few-shot Dataset}
%Before using the proposed few-shot selection methods, 
The entity types in NER datasets are usually not equally distributed. %If we directly select few-shot examples from the original dataset, we are highly likely to select the most frequent entity types and overlook the less frequent ones.
If we directly select few-shot examples from the original dataset, we risk focusing only on the most common entity types while overlooking the less common ones. 
To address this issue, we uniformly stratify a certain number of few-shot examples based on their entity types and build a stratified few-shot dataset. 
%This step is crucial for FsPONER, because the entity types in the original NER datasets are not equally distributed. %If we simply select few-shot examples from the original dataset, some infrequent entity types might not be included. 

We start by sorting all entity types from the raw dataset based on their frequency. Subsequently, we iterate through the examples of each entity type, and sequentially select one example from the most frequent entity to the least frequent one. The stratified dataset fairly considers less frequent entities and enables LLMs to learn more detailed information from the few-shot examples.

Furthermore, a smaller few-shot dataset aligns with our practical considerations, because in real-world projects, data annotation is expensive and only hundreds of annotated data are available. To simulate such low-resource scenarios, we limit the size of few-shot dataset to 300 examples in the experiments.

%Besides the few-shot dataset, we can integrate stratification into the three few-shot selection methods additionally, ensuring that the prompt contains at least one example of each entity type. If the allowed number of few-shot examples in a prompt is smaller than the number of different entity types, the samples will be sequentially drawn from the most frequent entity type to the least frequent one. We compare the stratified selection methods with their original versions in subsection \ref{stratified_selection}

\subsection{Few-shot Selection Methods}
\label{sec: methodology_llm_based_methods}
We illustrate FsPONER with three variants with different few-shot selection methods: random sampling, TF-IDF vectors, and a combination of both. % respectively to select semantically close examples for the prompt. 

\subsubsection{Selecting few-shot examples randomly}
Random sampling is an intuitive solution, in which samples are drawn from the few-shot dataset randomly and employed as few-shot examples directly. For each input sentence, the selected samples are non-repetitive, but they can be selected as few-shot examples for other input sentences repeatedly. While this approach is straightforward to implement, it overlooks the semantic relation between the sentences. Therefore, %we incorporate sentence encoders to attain sentence embedding or measure similarities between the computed TF-IDF vectors in the other two methods, aiming to select more fitting examples. 
we incorporate TF-IDF into the other two methods to filter more fitting examples.

\subsubsection{Selecting few-shot examples based on TF-IDF vectors}
As a weighting factor, Term Frequency–Inverse Document Frequency (TF-IDF) measures the significance of a word to a document in a corpus and finds wide applications in information retrieval and text mining tasks. 
\begin{figure}[h!]
    \includegraphics[width=0.8\linewidth]{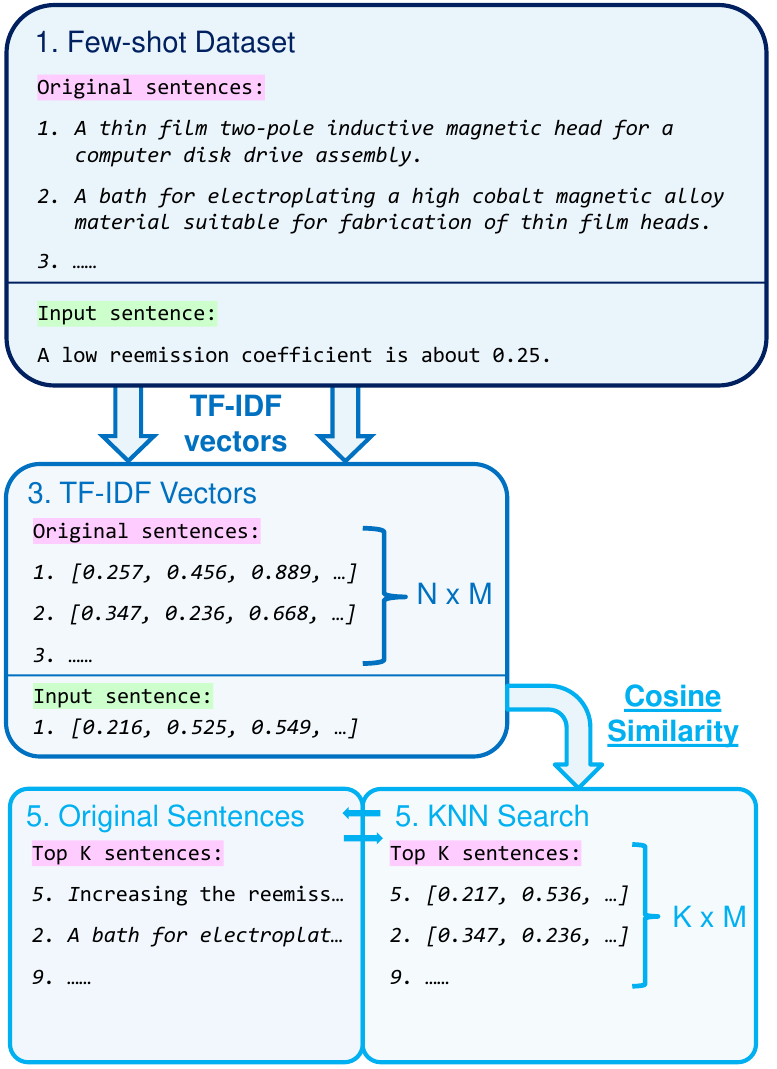}
    \centering
    \caption{The few-shot selection process based on TF-IDF vectors.
    }
    \vspace{0.5cm}
    \label{method_2_sentence_embeddings}
\end{figure}

%As illustrated in \ref{method_2_sentence_embeddings}, we stored the transformed embedding vectors in a \(N \cdot D\) matrix, i.e.~the number of examples $N$ times the dimension of embedding space $D$. We calculate the cosine similarity between two sentence embedding vectors and identify the $k$ nearest few-shot examples for each input sentence as few-shot examples. 

%The principle of the third method is similar to the second one, but we select few-shot examples based on the computed Term Frequency–Inverse Document Frequency (TF-IDF) vectors. 

%As demonstrated in \ref{method_2_sentence_embeddings}, 
The TF-IDF-based method shares a similar structure with GPT-NER~\cite{wang2023gpt}, but we substitute the sentence encoder with a TF-IDF transformer, converting both input sentences and few-shot examples into TF-IDF vectors. %, and thereby few-shot examples are selected based on the computed TF-IDF vectors rather than sentence embedding-vectors.. 
As illustrated in Fig.~\ref{method_2_sentence_embeddings}, we store the transformed TF-IDF vectors in an \(N \times M\) matrix, where $N$ is the number of few-shot examples and $M$ is the quantity of individual words within this corpus. We calculate the cosine similarity between two TF-IDF vectors and identify the $K$ nearest few-shot examples for each input sentence as few-shot examples. 
%Subsequently, we find the K nearest examples for the input sentence based on the cosine similarity and use them as few-shot examples. The matrix of TF-IDF vectors has a shape of \(N \cdot M\),  

\subsubsection{Selecting few-shot examples based on TF-IDF and random sampling}
Assuming that NER datasets exhibit a normal distribution of entities, we suggest the third method by integrating random sampling into the TF-IDF-based selection. Our goal is to ensure that the distribution of entities in the selected few-shot examples aligns with the overall few-shot dataset. 
For this purpose, we alternatively select few-shot examples from two sets of examples created through either random sampling or TF-IDF vectors until we reach the desired quantity.

\subsection{Prompt Structure}
Using the few-shot examples selected by FsPONER, we craft the prompt for domain-specific NER use cases. 
\begin{figure}[h!]
    \vspace{-0cm}
    \includegraphics[width=0.99\linewidth]{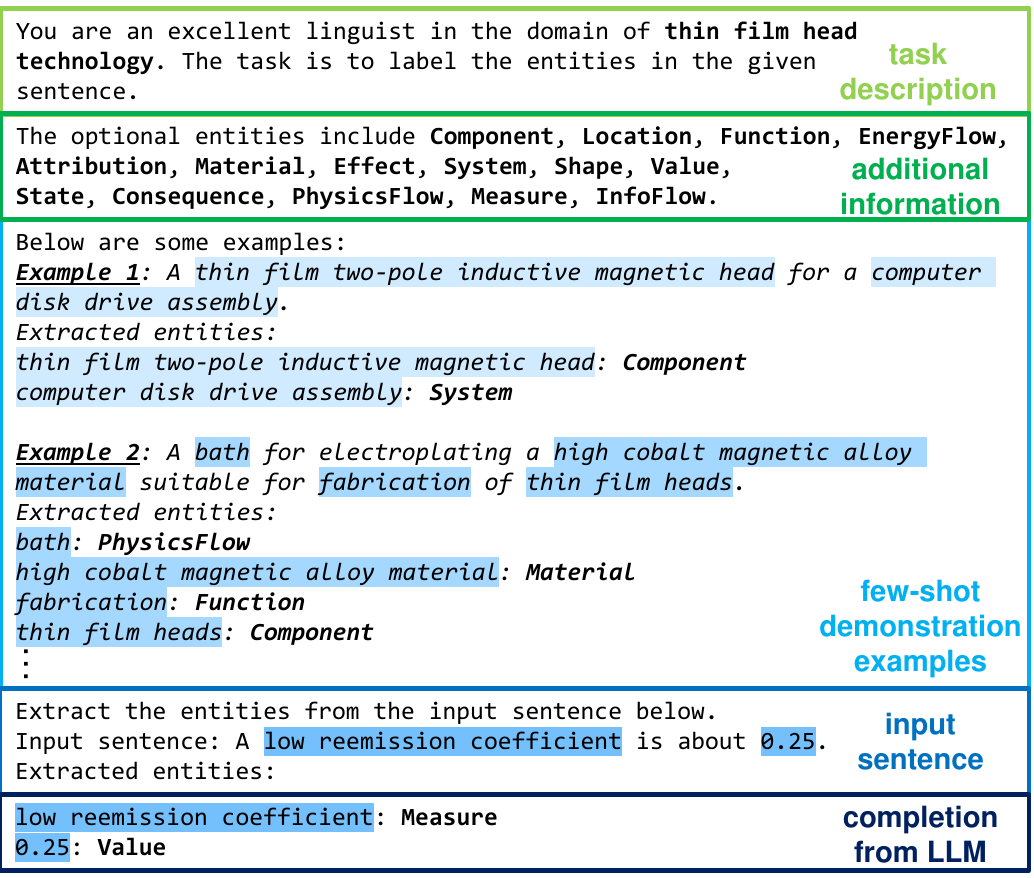}
    \centering
    \caption{The prompt structure for domain-specific NER tasks.}
    \vspace{0.55cm}
    \label{prompt_construction}
\end{figure}

Fig.~\ref{prompt_construction} illustrates the prompt structure, which consists of a concise task description, a paragraph of additional information, multiple few-shot examples, and the input sentence, from which LLMs identify the named entities. In the task description, we explain the NER task briefly and clearly define the dataset's domain. This allows the LLM to exploit the domain-specific knowledge acquired during the pre-training stage. Subsequently, we enumerate all pre-defined named entities as additional information to constrain the entity types for the LLM. In the third block, we add the selected few-shot examples to the prompt. %In \autoref{prompt_construction}, we highlight the words that should be extracted in these examples. 
The LLM learns from these examples and thereby extracts the entities more accurately during the inference stage. %, thus understanding the specific NER scenario better and generating the completion in the same format as demonstrated in few-shot examples. 
After the few-shot demonstration, we finally add the formal input sentence to the prompt and perform inference based on the given information. 
\section{Fine-tuning}
As LLM-based prompting becomes increasingly popular, there is a growing interest in the research community to evaluate their performance against fine-tuned language models. Previous works~\cite{wang2023gpt} have shown that general-purpose LLMs can achieve comparable performance to fine-tuned models in standard benchmarks. However, their efficacy in domain-specific scenarios has never been studied.
To address this gap, we fine-tune BERT and LLaMA 2-chat on three industrial datasets and compare the results with LLMs using FsPONER. We outline the procedure of fine-tuning LLaMA 2-chat 7B, with a focus on data pre-processing and low-rank adaptation~\cite{hu2021lora} to reduce GPU requirements.

%\subsection{Data preprocessing}
\subsection{Data Preprocessing}
\begin{figure}[!h]
    \vspace{-0cm}
    \includegraphics[width=0.88\linewidth]{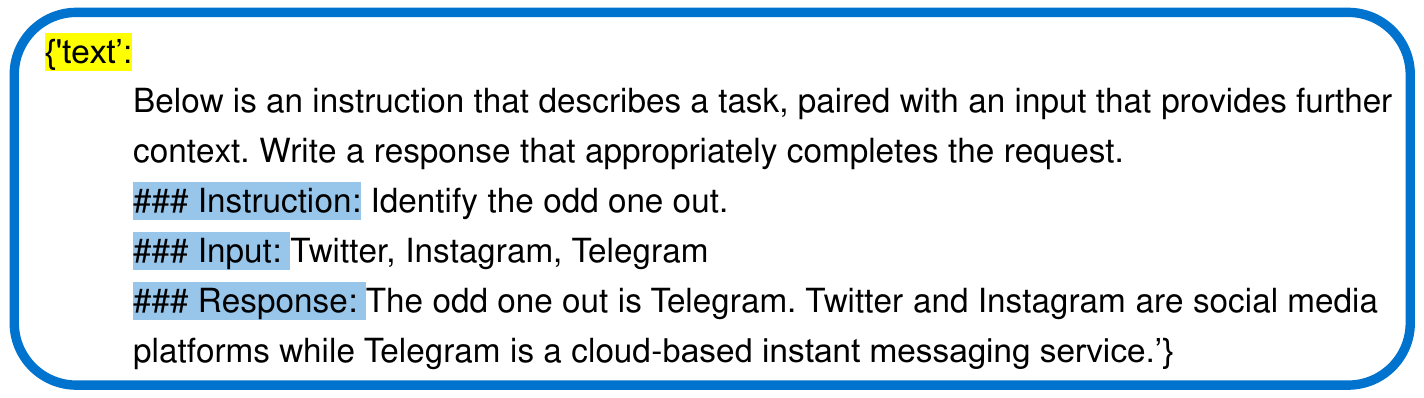}
    \centering
    \vspace{-0.1cm}
    \caption{A pre-processed prompt-completion pair for fine-tuning.}
    \vspace{0.3cm}
    \label{instruction_finetune_dataset}
\end{figure}

An unprocessed instruction dataset comprises instruction, input, and output columns. The instruction describes the task. The input provides further context of this task. The response represents the standard answer that LLMs should generate. As shown in Fig.~\ref{instruction_finetune_dataset}, we place these columns side by side to create a set of prompt-completion pairs for instruction fine-tuning, accompanied by an introductory explanation at the beginning to elucidate their respective roles. Throughout the training process, the LLM learns the statistical distribution of prompt-completion pairs, thereby advancing its understanding in the specific domain.
%\ref{data_preprocessing_instruction_fine-tuning_results} demonstrates an example of a pre-processed prompt-completion pair for instruction fine-tuning.

\subsection{Low-rank Adaptation}
% Training the LLaMA 2-chat 7B model in full precision is infeasible with our available GPU resources. To fine-tune the model more efficiently, we leverage Low-Rank Adaptation (LoRA)~\cite{hu2021lora} to reduce the hardware requirement on GPU memory while simultaneously maintaining the on-par performance. 
% % \begin{figure}[!h]
% %     \vspace{-0cm}
% %     \includegraphics[width=0.66\linewidth]{figures/lora.pdf}
% %     \centering
% %     \caption{The principle of LoRA.}
% %     \vspace{-0cm}
% %     \label{methodology_peft_illustration}
% % \end{figure}

\begin{figure}[!h]
    \vspace{-0cm}
    \includegraphics[width=0.64\linewidth]{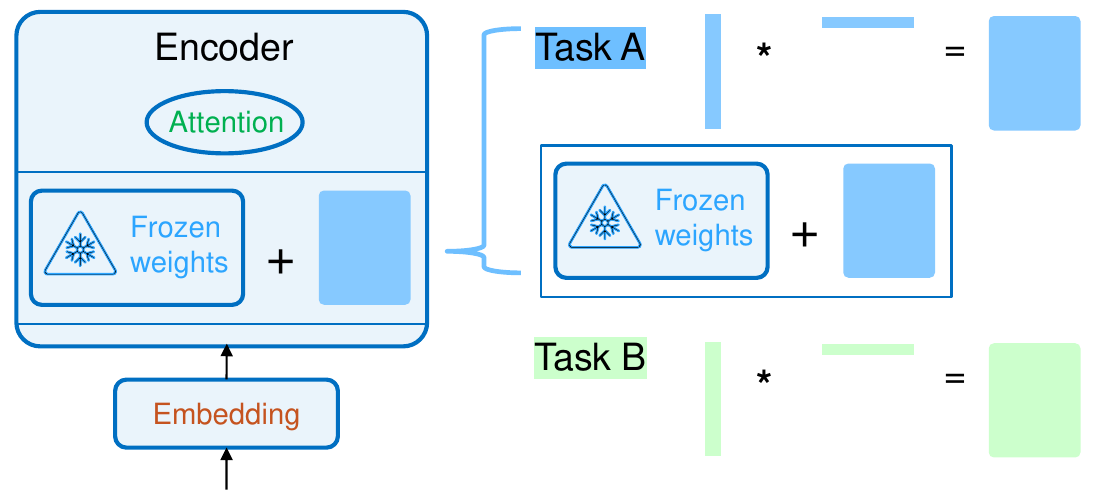}
    \centering
    \vspace{-0.2cm}
    \caption{The principle of LoRA.}
    \vspace{0.3cm}
    \label{methodology_peft_illustration}
\end{figure}

% In a fine-tuning process, all parameters of a LLM are updated. LoRA reduces the parameters to be trained by freezing all original model parameters and injecting a pair of low-rank decomposition matrices. We only update the two decomposition matrices in training. For different downstream tasks or datasets, we train different decomposition matrices and then add them to the original weights to update the values. The memory required to store these matrices is much smaller than training the entire model, which makes the entire fine-tuning process more accessible. 
Training the LLaMA 2-chat 7B model in full precision is infeasible with our available GPU resources. We leverage Low-Rank Adaptation (LoRA)~\cite{hu2021lora} in Fig.~\ref{methodology_peft_illustration} to reduce the hardware requirement on GPU memory while simultaneously maintaining the on-par performance. LoRA reduces the parameters to be trained by freezing all original model parameters and injecting a pair of low-rank decomposition matrices. We only update the two decomposition matrices in training. For different downstream tasks or datasets, we train different decomposition matrices and then add them to the original weights to update the values. The memory required to store these matrices is much smaller than training the entire model, which makes the entire fine-tuning process more accessible. 

%Researchers have found that applying LoRA solely to the self-attention layers of a model is often sufficient for a task, which achieves adequate performance gains. While using LoRA on feed-forward layers is also feasible, most of the parameters of LLMs are in attention layers and we attain the biggest savings in trainable parameters by applying LoRA to them. 

\subsection{BERT-based Fine-tuning}
%\subsection{Data Preprocessing for BERT-based Fine-tuning}

BERT variants find wide applications in diverse downstream tasks and many of them still hold the state-of-the-art performance on standard benchmarks. Therefore, we include BERT as a reference and fine-tune it for the considered NER datasets in a supervised fashion. 

\begin{figure}[!h]
    \vspace{-0.1cm}
\includegraphics[width=0.55\linewidth]{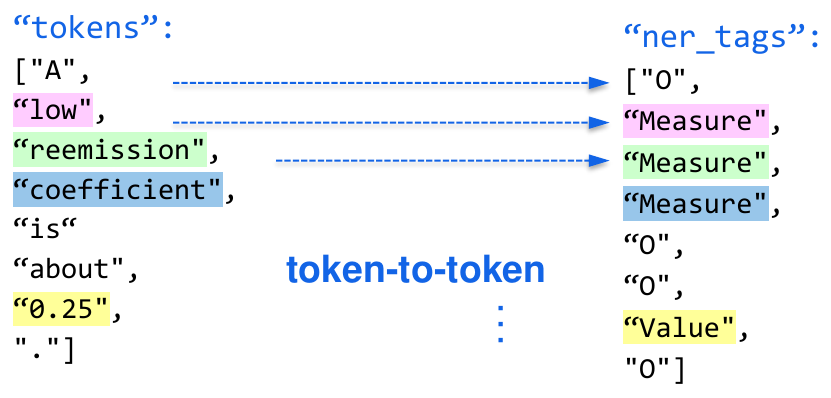}
    \centering
%\includegraphics[width=0.43\linewidth]{figures/methodology/BERT-NER.pdf}
    %\centering
    \vspace{-0.1cm}
    \caption{The BERT model formulates NER as a token-to-token task.}
    \vspace{0.8cm}
    \label{data_preprocessing_bert}
\end{figure}

As an encoder model, BERT approaches NER as a sequence labeling task and the data preprocessing differs from LLaMA 2-chat. As illustrated in Fig.~\ref{data_preprocessing_bert}, we convert the raw data into token-to-token JSON format. During the fine-tuning process, BERT learns to associate each individual word token with the correct entity type.

\section{Experiments}
%As proved by the previous studies in Chapter \ref{related_work}, 
Regarding LLM-based prompting, multiple influencing factors, such as the order, quantity, and quality of few-shot examples, can influence the performance of LLMs. Considering these factors, some existing works~\cite{wang2023gpt, xu2023large} optimize the few-shot examples for the prompt, progressively advancing the performance of LLMs on general NER benchmarks. However, none of them investigates the efficiency of few-shot prompting on domain-specific NER scenarios.
%Meanwhile, the research community shows growing interest in comparing the performance of fine-tuning and prompting in specific domains.
To answer the question, %we exemplify the performance of FsPONER on three NER datasets in industrial manufacturing and maintenance. Furthermore, 
we integrate FsPONER into four LLMs and compare the performance with fine-tuned BERT and LLaMA 2-chat 7B on three NER datasets in industrial manufacturing and maintenance, %investigating the advantages of both fine-tuning and prompting in the considered domain-specific scenarios.
exemplifying the advantages and drawbacks of few-shot prompting in the considered domain-specific scenarios.

\subsection{Experimental setting}

%\subsubsection{Selected LLMs}
\textbf{Selected LLMs:} We have experimented with 4 LLMs to investigate the efficiency of FsPONER. Table \ref{tab:LLM_comparison} provides an overview of these models based on their sizes, context window length, training data volume, allowed input modalities, and openness to the public. The two large-scale GPT models -- GPT-3.5-turbo~\cite{brown2020language} and GPT-4-32K~\cite{openai2023gpt4} are only accessible with OpenAI API. Their performance stands for the forefront of existing LLMs, which allows us to exploit the full potential of FsPONER. Furthermore, we include LLaMA 2-chat~\cite{touvron2023llama2}, one of the open-source top-performing models released by Meta AI, and Vicuna~\cite{zheng2023judging}, which is instruction fine-tuned from LLaMA 2-chat 13B on 125K instructional conversations generated by human beings. Alongside the four LLMs, we consider BERT~\cite{devlin2019bert} %for its superior performance in many downstream tasks 
and evaluate it as a reference. Since it does not demonstrate instruction-following abilities, we fine-tune it for NER tasks in a supervised fashion. 
\begin{table}[htpb]
  \caption[Example table]{Basic information of the selected LLMs}
  \vspace{0.4cm}
  \label{tab:LLM_comparison}
  \vspace{-0.1cm}
  \scalebox{0.75}{
      \begin{tabular}{l l l l l}
        \toprule
        \textbf{Models} & \textbf{GPT-3.5-turbo}&\textbf{GPT-4-32k}&\textbf{LLaMA 2-chat}&\textbf{Vicuna V1.5}\\%&\textbf{BERT}\\
        \midrule
        Model size & 175B & - & 7B/13B/70B& 7B/13B \\%&110M/340M \\
        (parameters)& & & & 
        \medskip\\
        Context window &4096&32768&4096&4096\\%&512\\
        length (tokens)
        \medskip\\
        Training data& 300B tokens& - & 2T tokens&2T tokens +\\%& 3.3B words\\
        volume & for GPT-3& & &125K samples 
        \medskip\\
        % Training process&-%\footnote{RLHF stands for reinforcement learning from human feedback, first proposed by christiano et al.~\cite{christiano2017deep}}
        % &-&RLHF%\footnote{RLHF stands for Reinforcement Learning from Human Feedback.}
        % &Instruction&pre-trained\\
        % & & & &fine-tuned& 
        % \medskip\\
        Modalities & text & text/image & text & text \\%& text \\
        & &(input) &  &    
        \medskip\\
        Openness & closed & closed & open & open \\%& open \\
        \bottomrule
      \end{tabular} 
  }  
  \centering
\end{table}

\textbf{Selected datasets:} In the experiments we consider the following three publicly available real-world datasets from industrial manufacturing and maintenance -- the thin-film head technology dataset~\cite{film2020chen}, the assembly instruction dataset~\cite{assembly2017costa}, and the manufacturing dataset~\cite{fabner2022kumar}. We present their basic information in Table \ref{tab:dataset_comparison}.

\begin{table}[htpb]
  \caption[Example table]{Basic information of the three NER datasets}
  \vspace{0.4cm}
  \label{tab:dataset_comparison}
  \vspace{-0.1cm}
  \scalebox{0.75}{
      \begin{tabular}{l l l l l l}
        \toprule
        \textbf{Datasets} & \textbf{Number}&\textbf{Number}&\textbf{Domain}&\textbf{State-of-}&\textbf{Examples}\\
         & \textbf{of words}&\textbf{of types}& &\textbf{the-art}&\textbf{of types}\\
         & & & &\textbf{(F1 score)}& \\
        \midrule
        FabNER&350,000+&12&Manufac-&92\%&MATE(Material),\\
         & & & turing& &PRO(Properties),\\
         & & & & &ENAT(Enabling\\
         & & & & &technology),\\
         & & & & &...\\
        \vspace{-0.2cm} 
        \smallskip\\
        Thin-film&92,000+&17&Hard-disk&78.2\%&Component,\\
         head& & & & &Function,\\
         technology& & & & &PhysicsFlow,\\
         & & & & &EnergyFlow,\\
         & & & & &...\\
        \vspace{-0.2cm} 
        \smallskip\\
        Assembly&22,000+&9&Assembly&84.69\% &PART(parts),\\
        NER& & & & &OPER(operations),\\
         & & & & &TOOL(tools),\\
         & & & & &QTY(Quality),\\
         & & & & &...\\        
        \bottomrule
      \end{tabular} 
  }  
  \vspace{-0cm}
  \centering
\end{table}
% \begin{figure}[!h]
%     \vspace{-0cm}
%     \includegraphics[width=0.99\linewidth]{ecai-template/figures/dataset.pdf}
%     \centering
%     \caption{The basic information of three industrial datasets.}
%     \vspace{-0cm}
%     \label{basic_dataset_info}
% \end{figure}

%While FabNER manufacturing and thin-film technology datasets contain thousands of examples, Assembly NER consists of 800  

\textbf{Few-shot dataset:} For real-world use cases, data annotation proves to be a cost-intensive endeavor. Utilizing limited high-quality data resources to enhance the performance of LLMs
is a sensible and realistic strategy. In line with this practical consideration, we create a stratified few-shot dataset of 300 samples. From our empirical practice, this is an applicable volume that keeps data annotation expenses within a manageable bound. When constructing the prompt, few-shot examples will be selected from this dataset.

\subsection{Results of FsPONER}
\label{stratified_selection}
%Why we do this experiment? its connection to the motivation?
We compare the three few-shot selection methods proposed for FsPONER with GPT-NER~\cite{wang2023gpt} as we increase the number of few-shot examples in the prompt and attempt to identify the top-performing setting for FsPONER.

Due to the massive scale of conducted experiments and some repetitive results across the three datasets, we exclusively demonstrate the evaluation results on the thin-film technology dataset and perform in-depth analysis for the selected models. %For the rest of selected models, we show their evaluation results on thin-film technology dataset, which has covered most of the main research findings on these models. 
For readers who are interested, the evaluation results on the other two datasets are available in supplementary documents~\cite{tang2024}. 

\begin{figure}[!h]
    \vspace{-0cm}
    \includegraphics[width=0.8\linewidth]{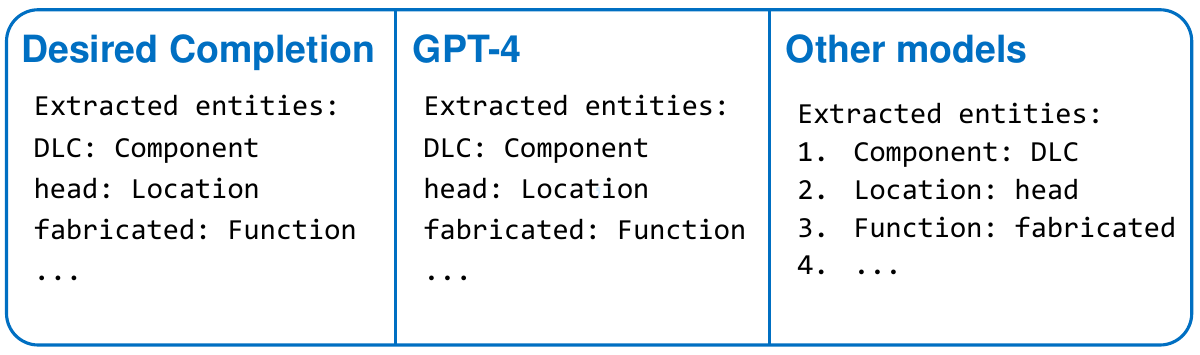}
    \centering
    \vspace{-0.15cm}
    \caption{The generated completions in zero-shot setting.}
    \vspace{0.5cm}
    \label{zeroshot_results}
\end{figure}

Fig.~\ref{poner_thin_film} illustrates the evaluation results on the thin-film head technology dataset, with few-shot selection methods on the horizontal axis and F1 score on the vertical axis. Multiple performance-changing rules of LLMs are evident. All evaluated models advance their performance across all few-shot selection methods as the number of few-shot examples increases. With an extended context window, GPT-4 leads the performance and achieves the optimal F1 score of 68.76\% by integrating 80 few-shot examples selected by FsPONER with TF-IDF. Furthermore, GPT-4 demonstrates exceptional capability in correctly understanding instructions. In the zero-shot evaluation, only GPT-4 can strictly adhere to the format described in text when generating entity types. The other LLMs may modify the format spontaneously, e.g.~adding a serial number or placing the corresponding entity types before the original words, as illustrated in Fig.~\ref{zeroshot_results}. Additional steps are required to process these generated completions.

For the other models, due to the limited context window of 4096 tokens, we are not allowed to add more than 20 few-shot examples to the prompt. % and have reached their best achievable performance. 
When using Fs-PONER with TF-IDF, Text-Davinci-003 from the GPT-3-Turbo family demonstrates the optimal performance, reaching a weighted F1 score over 66{\%}, which is 2.7\% lower than GPT-4-32K. However, if we feed both GPT models with 20 few-shot examples, Text-Davinci-003 leads the F1 score by 0.5\%.

\begin{figure}[!h]
    \vspace{-0cm}
    \hspace{-0.3cm} 
    \includegraphics[width=1.0\linewidth]{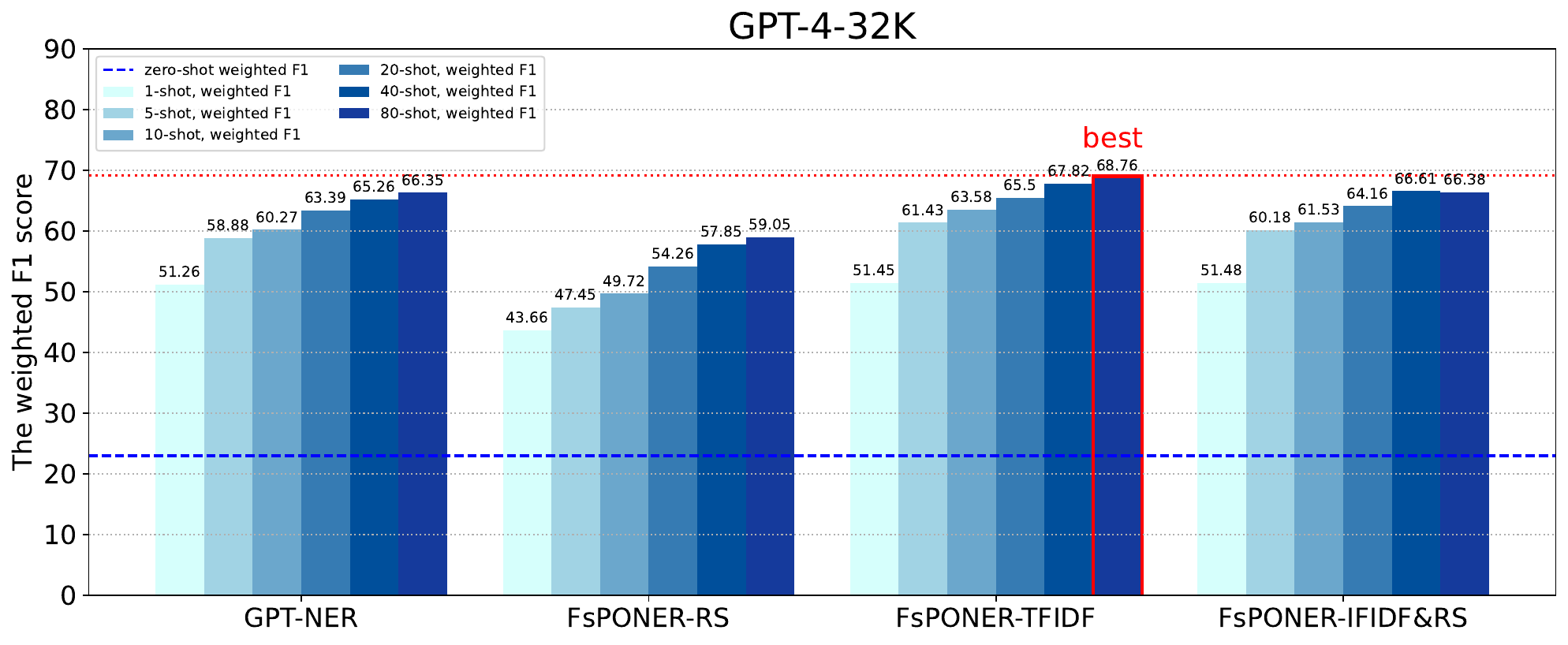}
    %\centering
    \includegraphics[width=1.0\linewidth]{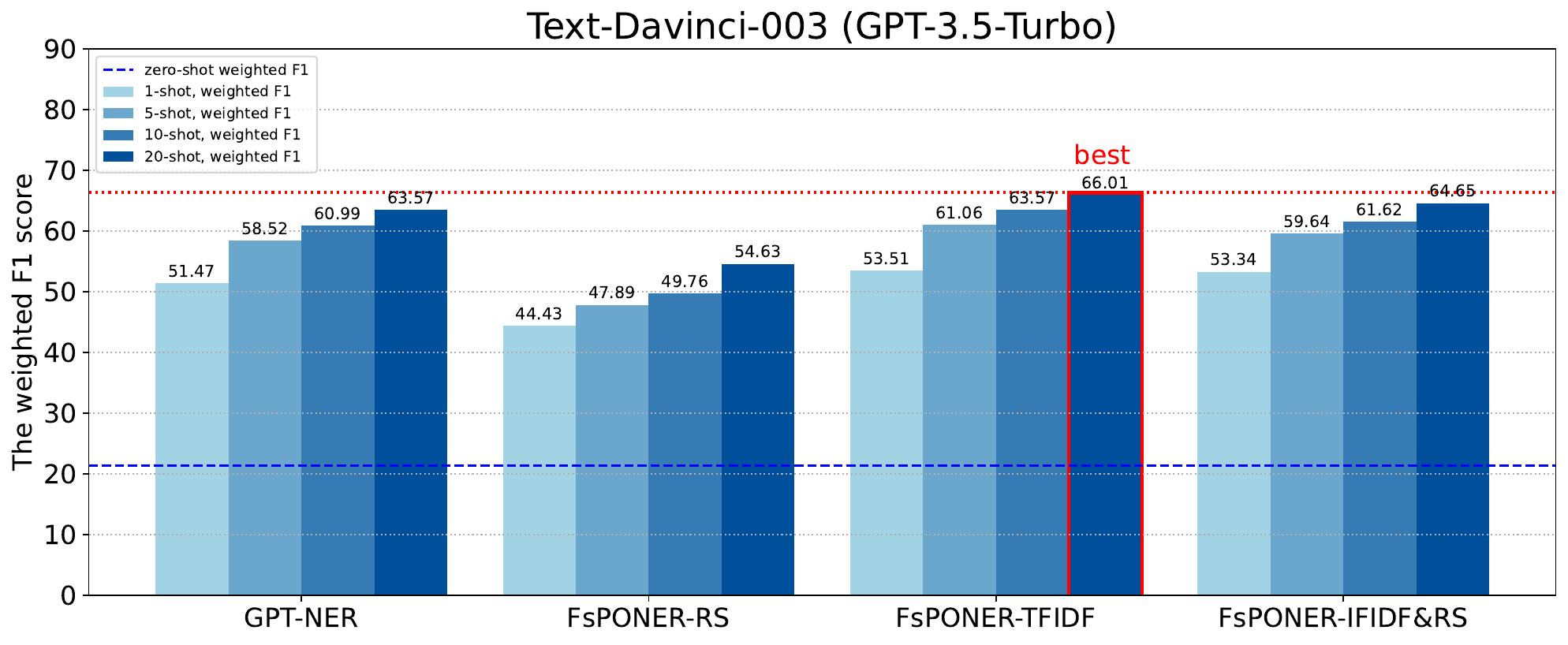}
    %\centering
    \includegraphics[width=1.0\linewidth]{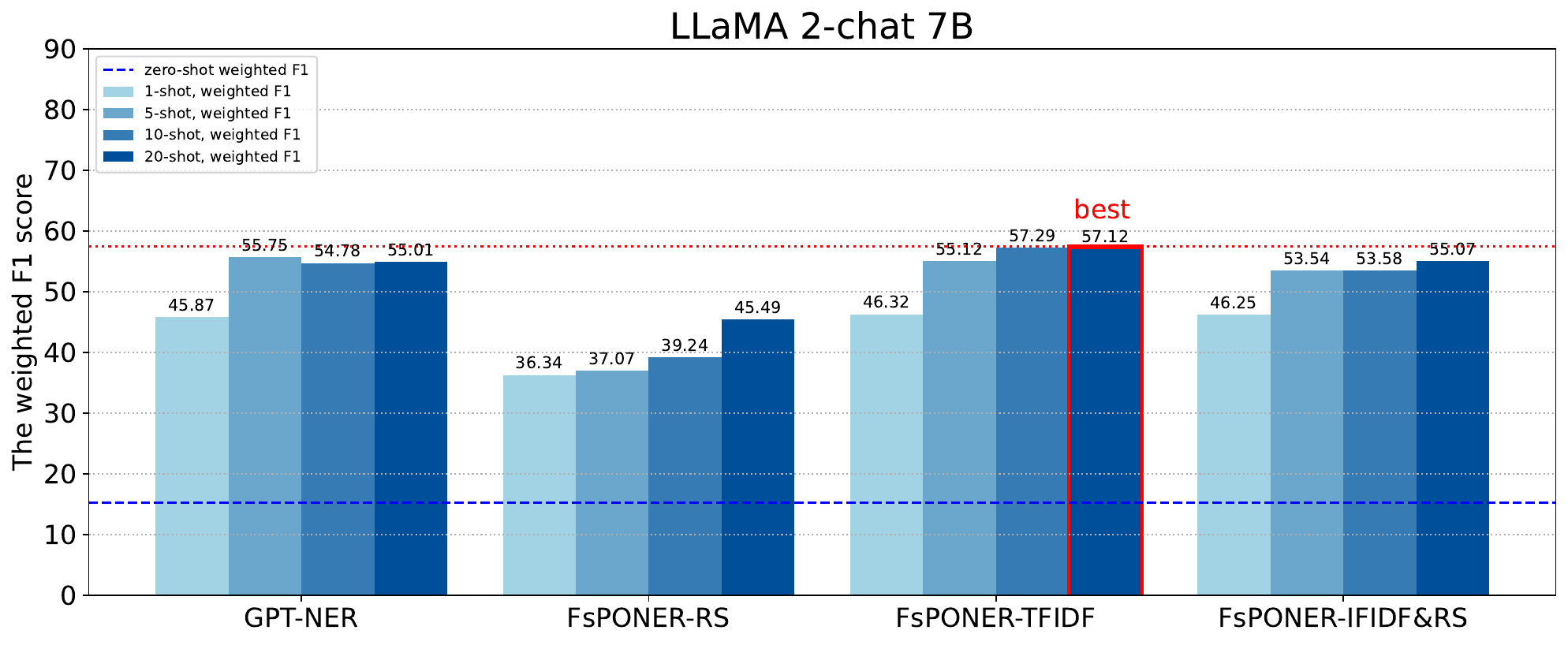}
    %\centering
    \includegraphics[width=1.0\linewidth]{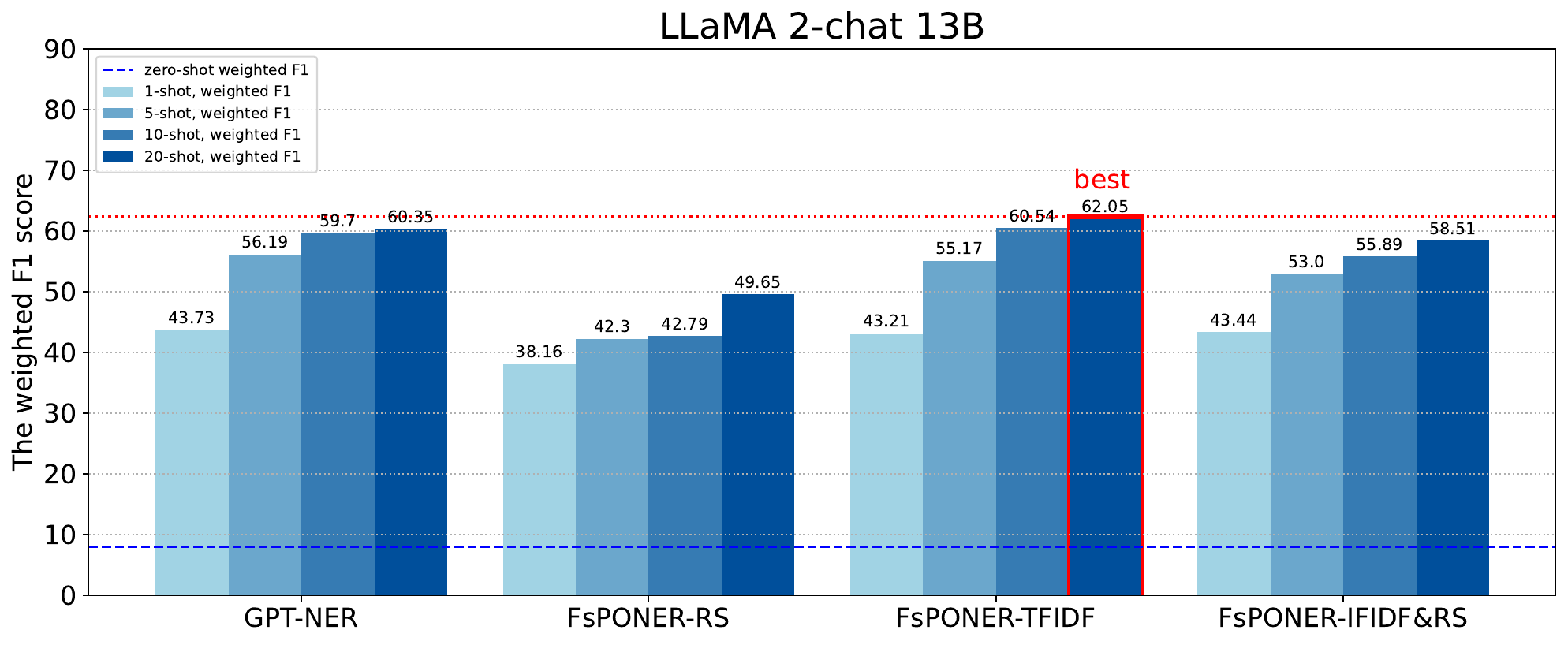}
    \includegraphics[width=1.0\linewidth]{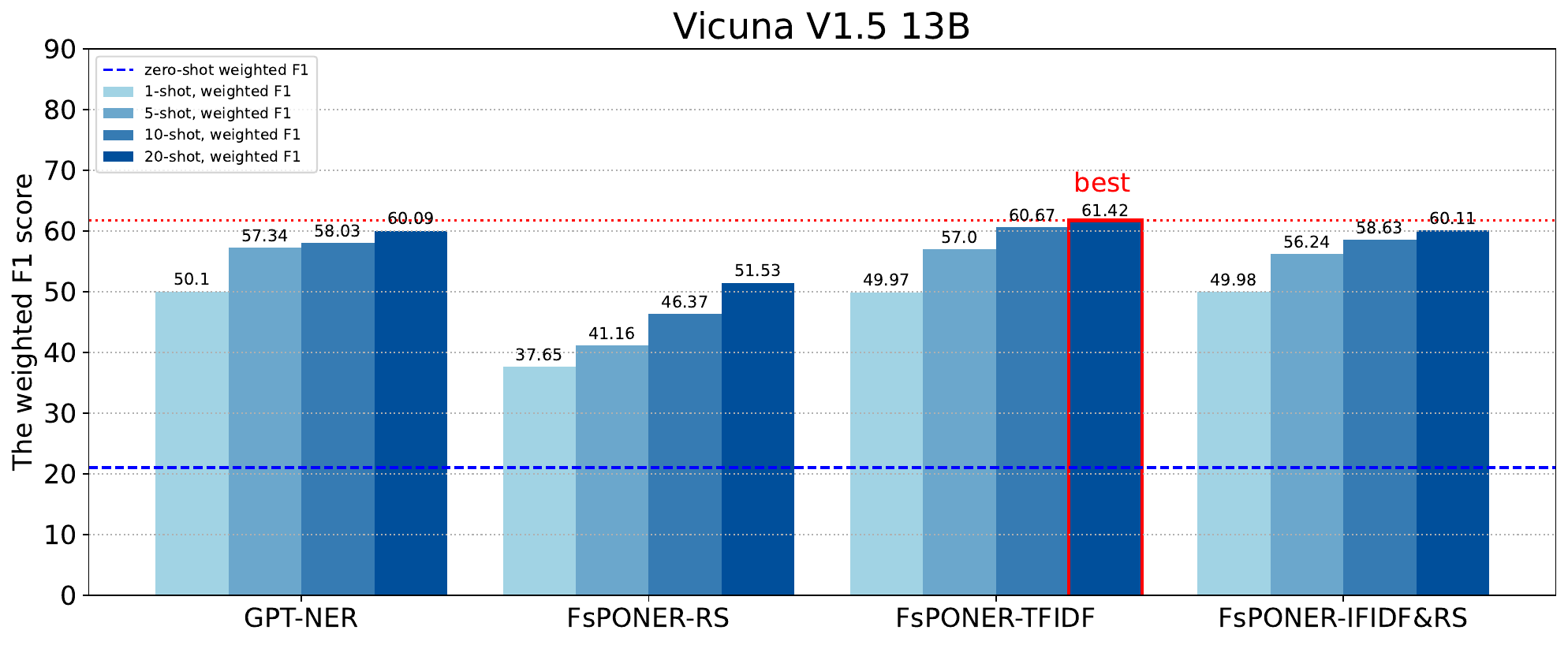}
    %\centering
    \vspace{-0.3cm}
    \caption{The evaluation results on the thin-film technology dataset. \hspace{1cm}(RS refers to random sampling.)%(RS / TFIDF / TFIDF\&RS refer to FsPONER with random sampling, TF-IDF vectors, and TF-IDF plus random sampling.)
    }
    \vspace{0.6cm}
    \label{poner_thin_film}
\end{figure}
Due to the limited model size, the performance of the selected open-source models falls behind the two GPT models. With respect to LLaMA 2-chat, the 13B version surpasses the 7B model in multiple facts. The performance of LLaMA 2-chat 7B hits a plateau once we have added 5 few-shot examples to the prompt. By contrast, the LLaMA 2-chat 13B demonstrates stronger ability to comprehend long context and the performance continues to improve as we increase the number of few-shot examples, reaching a F1 score of 62.05\%, which is 4.97\% higher than the 7B model when employing FsPONER with TF-IDF. %Considering the compact model size of LLaMA 2-chat 7B, 

Vicuna-V1.5-13B is instruction fine-tuned from LLaMA 2-chat 13B using 125K human-generated conversations, the overall NER performance is close to the fundamental model, reaching a weighted F1 score of 61.42{\%}. This observation indicates that fine-tuning on more generic conversational data does not enhance LLM's performance in a industrial domain. 

If we zoom in GPT-4-32K, the performance has not saturated. When we double the number of few-shot examples from 40 to 80, an improvement by around 1\% can still be observed in FsPONER with TF-IDF. Therefore, we integrate more few-shot examples, aiming to identify the maximum achievable performance of GPT-4-32k on the three considered NER datasets. 
\begin{figure}[!h]
    \vspace{-0cm}
    \hspace{-0.3cm}
    \includegraphics[width=0.52\linewidth]{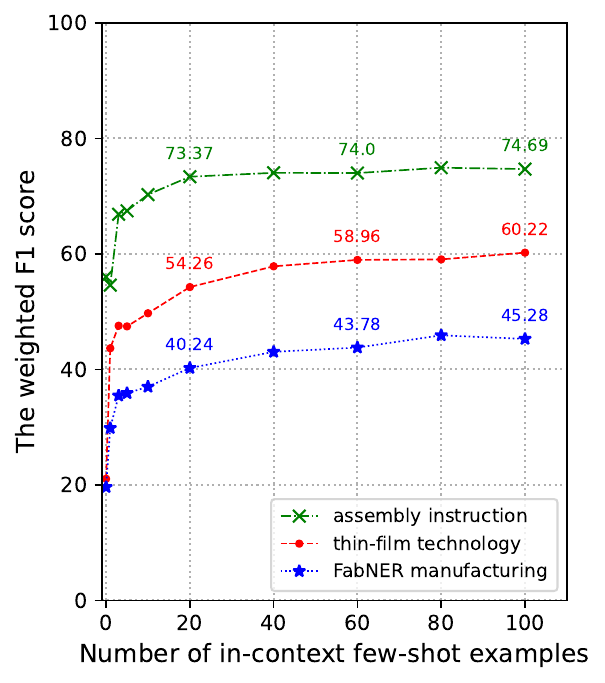}
    \hspace{-0.3cm}
    \includegraphics[width=0.52\linewidth]{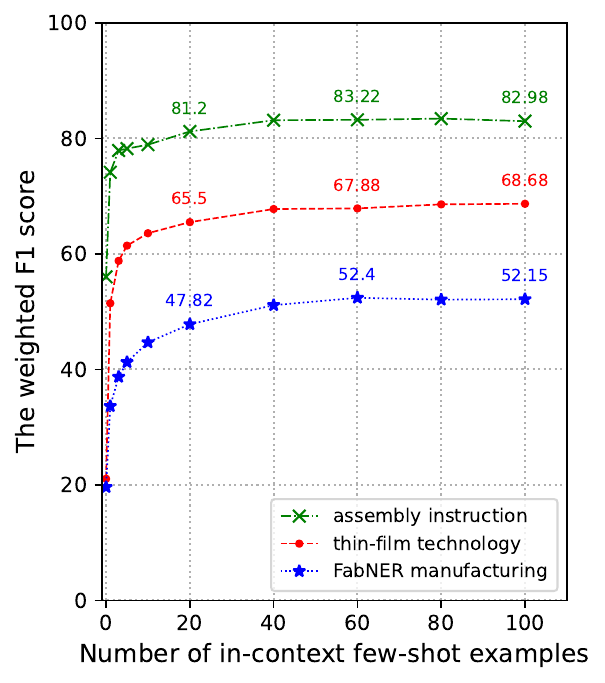}
    \hspace{0.3cm}
    \vspace{-0.4cm}
    \caption{The F1 score of GPT-4-32K on three NER datasets with a growing number of few-shot examples selected randomly (left) or using TF-IDF vectors (right).}
    \vspace{0.66cm}
    \label{performance_saturation}
\end{figure}

According to the previous experiments, FsPONER with TF-IDF achieves the top-notch performance among all few-shot selection methods. In light of this conclusion, we analyze the advantages of TF-IDF against random sampling %to investigate its advantages in detail 
and illustrate the F1 score evolution as we increase the number of few-shot examples to 100 in Fig.~\ref{performance_saturation}. 

For both few-shot selection methods, F1 scores improve dramatically in the first 20 few-shot examples and the trend slows down after reaching 40 examples. The NER performance with 60 few-shot examples closely approaches the performance with 100 examples, showing a difference below 1.5\% for random sampling and 0.8\% for TF-IDF in all three datasets. After the performance curve gradually levels off, TF-IDF secures a higher F1 score. Furthermore, compared to the twisting curve resulting from the unpredictable quality of random examples, the performance advances coherently and consistently in TF-IDF. The saturation phenomenon of F1 score exhibits the performance plateau of prompting with more few-shot examples. We have reached the optimal performance of FsPONER.

\subsection{Results of Fine-tuning}
%Why we do this experiment? its connection to the motivation?
%Following the optimal setting of FsPONER, we continue the experiments and compare the LLM-based few-shot learning with the fine-tuned language models, investigating the advantages of both methods in the considered use cases. While the LLMs using FsPONER select examples from the few-shot dataset, Bert and LLaMA 2-chat are fine-tuned with the full training dataset in our experiments. 
Following the evaluation results in Fig.~\ref{poner_thin_film} and Fig.~\ref{performance_saturation}, %we standardize the top-performing few-shot learning configuration to FsPONER with TF-IDF utilizing 80 few-shot examples
we use FsPONER with TF-IDF for few-shot selection, integrating 80 few-shot examples in the prompt for each input sentence, and compare its performance with the fine-tuned BERT and LLaMA 2-chat 7B. While the LLMs using FsPONER select examples from the few-shot dataset, Bert and LLaMA 2-chat are fine-tuned with the full training dataset in our experiments. 

We include an unsupervised NER baseline additionally to measure the advanced performance of language models. %The idea originates from $k$NN-NER~\cite{wang2022k}. %, where the entity type of each word is decided based on the $k$ nearest tokens retrieved from the prior datastore. 
While the initial method~\cite{wang2022k} integrates the $k$NN algorithm into the vanilla BERT model as a complete system, we directly utilize sentence-BERT to obtain the embedding of each token. This adaptation enables us to avoid training BERT, which overlaps with the bert-based fine-tuning approach. 

\begin{figure}[!h]
    \vspace{-0cm}
    \includegraphics[width=1\linewidth]{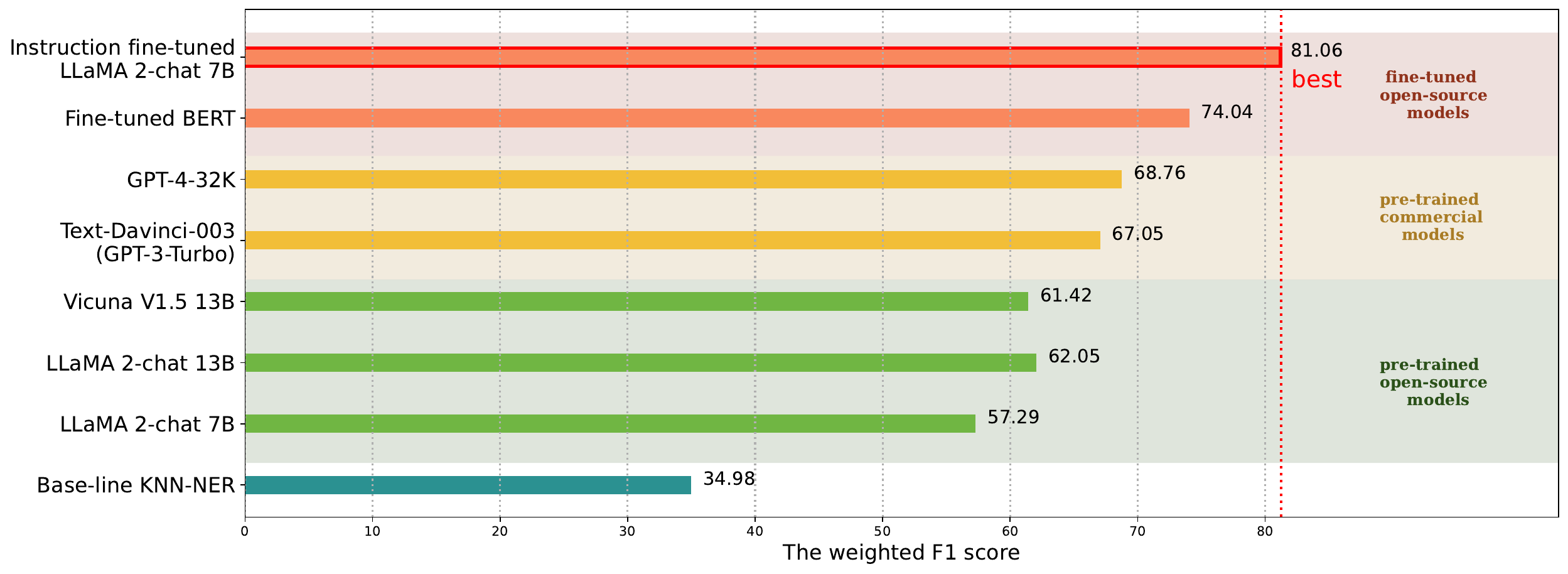}
    \centering
    \includegraphics[width=1\linewidth]{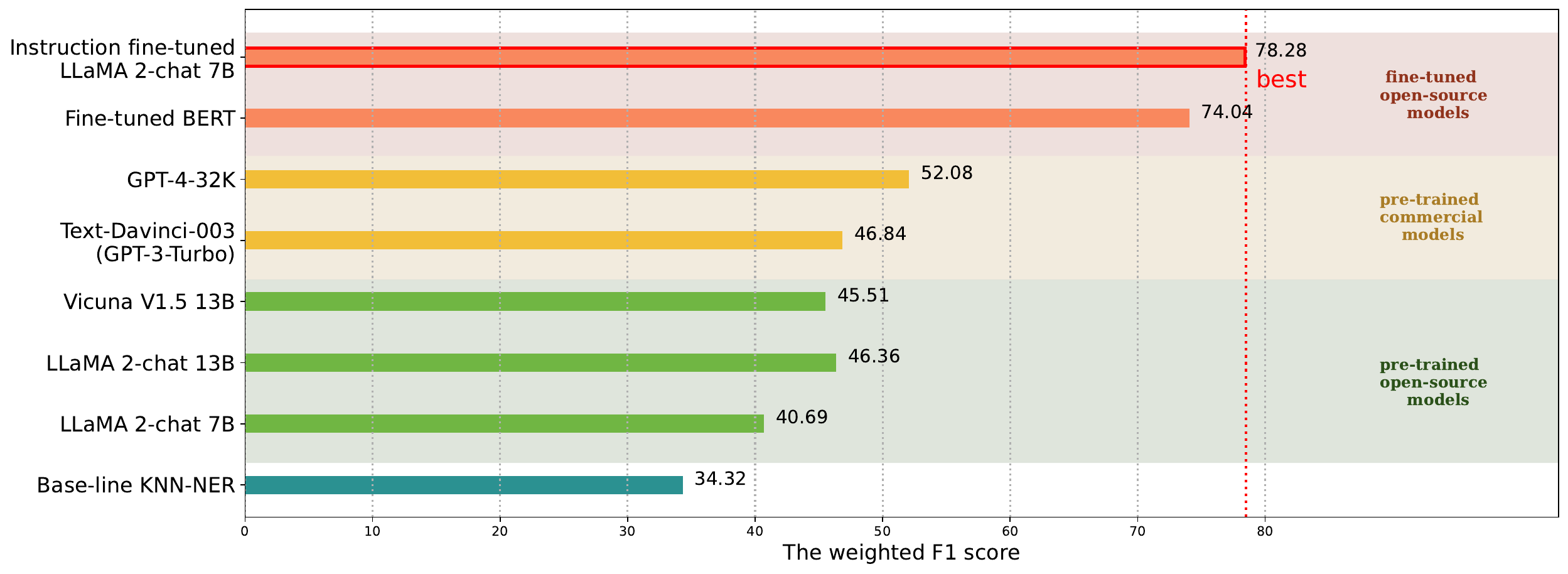}
    \centering
    \includegraphics[width=1\linewidth]{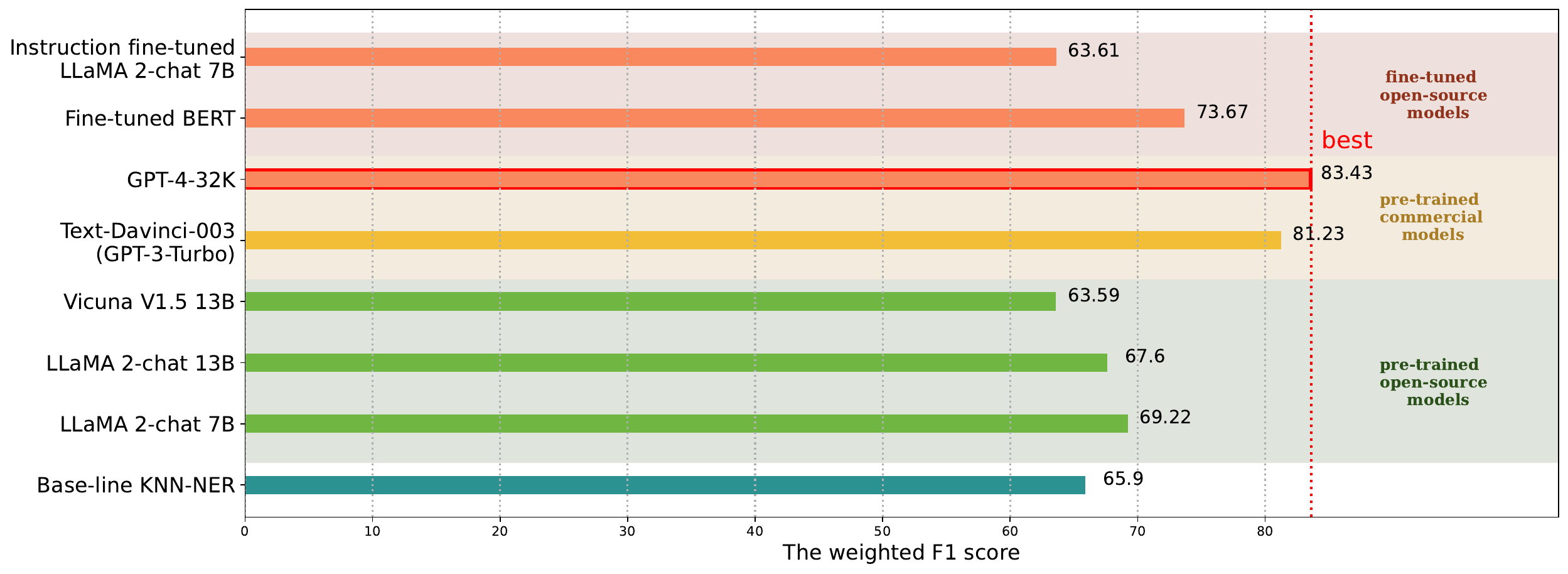}
    \centering
    
    \caption{Overal NER performance of different language models on the thin-film technology, FabNER manufacturing, and assemblyNER datasets (from top to bottom).}
    \vspace{0.6cm}
    \label{performance_overview}
\end{figure}

As illustrated in Fig.~\ref{performance_overview}, the fine-tuned models outperform the pre-trained LLMs using FsPONER in the two larger datasets. In the FabNER dataset, the fine-tuned LLaMA 2-chat 7B obtains 78.28\% and the fine-tuned BERT reaches 74.04\%, which leads all pre-trained models by more than 20\% in F1 score. We observe analogous results from the thin film technology dataset, where the fine-tuned LLaMA 2-chat 7B achieves the optimal performance of 81.06\%, higher than the state-of-the-art presented in Table \ref{tab:dataset_comparison}. 
However, in the assembly dataset, which consists of fewer data but with more generic entity types, the two GPT models using FsPONER surpass the fine-tuned models, reaching 83.43\% and 81.23\% in F1 score respectively. One explanation for this phenomenon is the data scarcity in assembly domain, preventing effective fine-tuning of BERT and LLaMA 2-chat 7B to attain their optimal performance.

%\subsection{Model size and performance}
In the experiments, we confine the source of few-shot examples to the stratified few-shot dataset while fine-tuning with the full dataset. %, which is unfair in terms of the applied data quantity. %Considering the difference of the applied data quantity, the comparison between them is unfair. 
Considering the difference in the quantity of applied data, we continue investigating FsPONER with a varying size of few-shot dataset and extend the data source to the full dataset.
\begin{figure}[!h]
    \vspace{-0cm}
    \hspace{-0.2cm}
    \includegraphics[width=1\linewidth]{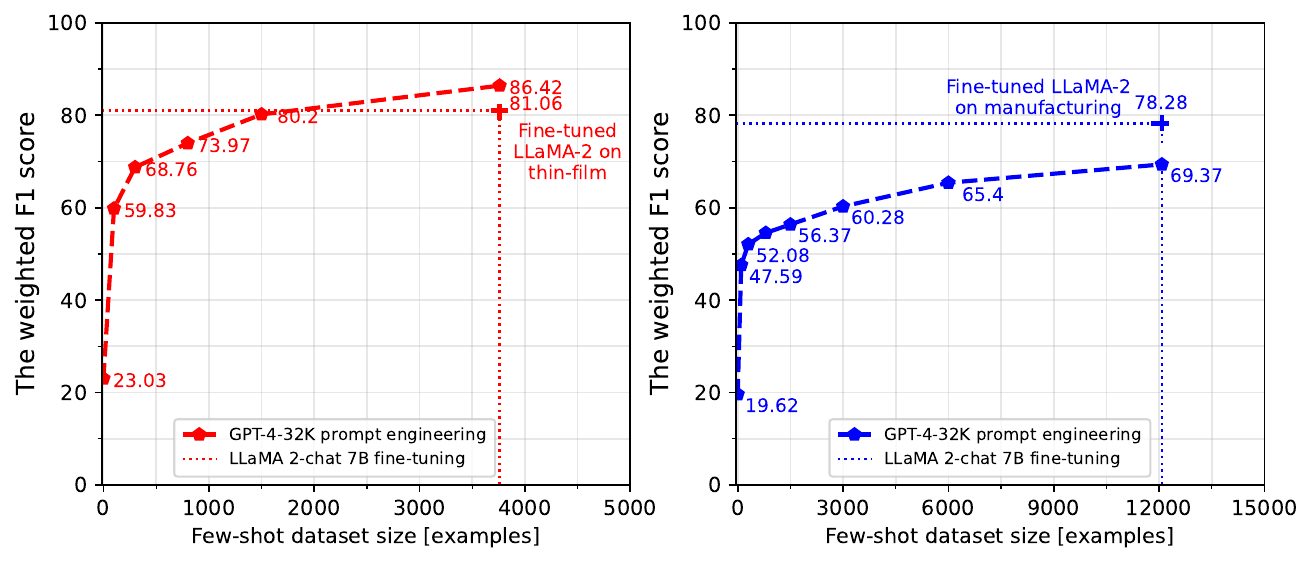}
    %\centering
    \vspace{-0.1cm}
    \caption{The F1 score of GPT-4 with an increasing size of few-shot dataset in thin-film technology (left) and manufacturing (right) domains.}    
    \vspace{0.4cm}
    \hspace{-1cm}
    \label{saturation_varing_development dataset}
\end{figure}
Fig.~\ref{saturation_varing_development dataset} illustrates the results, where we progressively expand the size of few-shot dataset from 100 examples to the full dataset. The assemblyNER dataset is not included, due to its limited data volume. For the thin-film technology and FabNER manufacturing datasets, the F1 score improves as the few-shot datasets scale up. When we utilize the full dataset, GPT-4 can achieve a F1 score of 86.42\% on the thin-film technology dataset, which surpasses the fine-tuned LLaMA 2-chat 7B by 5.36\%. In the FabNER dataset, GPT-4 advances the 52\% F1 score obtained with 300 few-shot examples to 69\% by utilizing the entire dataset, which is approximately 9\% lower than the fully fine-tuned LLaMA 2 chat. 

\subsection{Discussion}
The performed experiments demonstrate that data quantity, domain specificity, and the model capabilities significantly influence FsPONER's performance in domain-specific NER tasks. 
In a scenario with generic entity types, i.e.~in the assemblyNER dataset, FsPONER with a small set of high-quality data can outperform fine-tuning with the full dataset. However, in a specific domain with abundant data, e.g.~in the FabNER dataset, fine-tuning still leads the performance. For both fine-tuning and the FsPONER framework, larger models demonstrate more advanced performance in general, but they require massive training data and sufficient compute budget. We must consider the available GPU resources, the allowed training time, and the resulting cost. With a model of proper size and a well-suited method, we can achieve the optimal performance within a cost-effective setting.

\section{Conclusions and Future Work}
%In this work, we propose FsPONER, a LLM-based framework consisting of three few-shot selection methods based on random sampling, TF-IDF vectors, and a combination of both. We evaluate FsPONER on domain-specific NER scenarios, with a focus on industrial manufacturing and maintenance while using four LLMs, and then compare the evaluation results with fine-tuned BERT and LLaMA 2-chat. In the considered domain-specific NER scenarios, FsPONER with TF-IDF achieves the top-notch performance among all few-shot selection methods and it surpasses fine-tuned models by approximately 10\% in F1 score in certain industrial scenario with generic entity types. % by approximately 10\% in F1 score. Moreover, as we increase the quantity of few-shot examples in the prompt or augment the size of few-shot datasets, the performance of FsPONER continues to improve.

In this work, we proposed FsPONER, a few-shot selection framework for domain-specific NER tasks. In the considered NER scenarios, FsPONER with TF-IDF consistently demonstrates the top-notch performance compared to a general-purpose GPT-NER method and all other FsPONER variants. As we increase the quantity of few-shot examples in the prompt or expand the size of few-shot datasets, the performance of FsPONER continues to improve. Specifically, in an industrial manufacturing scenario with data scarcity, FsPONER with TF-IDF outperforms the fine-tuned models by approximately 10\% in F1 score.

As for future work, more strategies against hallucination are required to generate solid entities and avoid varying completion formats in Fig.~\ref{zeroshot_results}. Furthermore, the long inference time of LLMs necessitates an efficient solution, especially when dozens of few-shot examples are added to the prompt. Moreover, the leaderboard of LLMs is continuously updated. Many compact models have been specifically designed for information extraction and should be assessed for domain-specific NER tasks, which may have the potential to surpass the currently top-performing LLMs.

\bibliography{mybibfile}

\end{document}